\documentclass{article}

\usepackage[preprint]{neurips}

\usepackage[utf8]{inputenc} % allow utf-8 input
\usepackage[T1]{fontenc}    % use 8-bit T1 fonts
\usepackage{hyperref}       % hyperlinks
\usepackage{url}            % simple URL typesetting
\usepackage{booktabs}       % professional-quality tables
\usepackage{amsfonts}       % blackboard math symbols
\usepackage{nicefrac}       % compact symbols for 1/2, etc.
\usepackage{microtype}      % microtypography
\usepackage{xcolor}         % colors
\usepackage{algorithmic}
\usepackage{algorithm}
\usepackage{amsmath}
\usepackage{mathtools}
\usepackage{amsfonts}
\usepackage{subfig}
\usepackage{amssymb}
\usepackage{caption}
\usepackage{multicol}

\title{Anomaly Detection in Global Financial Markets with Graph Neural Networks and Nonextensive Entropy}

\author{
  Kleyton da Costa \\
  PUC-Rio, Rio de Janeiro, Brazil \\
  Holistic AI, London, United Kingdom\\
  \texttt{kcosta@inf.puc-rio.br} \\
  \texttt{kleyton.costa@holisticai.com} \\
}

\begin{document}

\maketitle

\begin{abstract}
	Anomaly detection is a challenging task, particularly in systems with many variables. Anomalies are outliers that statistically differ from the analyzed data and can arise from rare events, malfunctions, or system misuse. This study investigated the ability to detect anomalies in global financial markets through Graph Neural Networks (GNN) considering an uncertainty scenario measured by a nonextensive entropy. The main findings show that the complex structure of highly correlated assets decreases in a crisis, and the number of anomalies is statistically different for nonextensive entropy parameters considering before, during, and after crisis. 
\end{abstract}

\begin{multicols}{2}
% Introduction
\section{Introduction}

One of the main challenges in different fields is anomaly detection \cite{CHANDOLA2009ANONDET}, especially in systems influenced by high-dimensional data \cite{DENG2021GRAPHTS}. When these variables are observed in multivariate (or high-dimensional) time series data the task becomes even more difficult. Anomaly detection in high-dimensional data can be applied in the detection of financial fraud in bank transfers, monitoring of patients in intensive care units, monitoring of industrial processes, identification of outbreaks of infectious diseases, and traffic control on highways and railways.

Anomaly detection is related to identifying abnormal behaviors in a dataset. We consider outliers the data points that statistically differ from the data. These outliers can arise from at least three causes: (i) the occurrence of rare events (such as in the case of an economic crisis or pandemic), (ii) malfunction (such as in the case of a system for manufacturing products), or (iii) misuse of a system (such as in the case of attempted invasion or deviation of functionalities). Anomalies can be represented through various types of data, and in this study, our research interest is focused on networks.

The use of networks has gained prominence over the past decades for its ability to relate social, physical, biological, economic phenomena, etc. relatively simply. Networks can be mathematically represented through graphs in which a set of nodes relates through edges, creating information in a network. The nodes represent elements that are part of a certain set $V$ and the edges are the simple representation of the connections between the elements of $V$, defined as a set of edges $E$. When we combine the elements of $V$ and $E$, we have as a result a graph $G$. Thus, a given graph $G$ is defined by $G = (V_{i},E_{j}), ~ \forall ~ i = (1, 2, \dots, n) ~ \text{and} ~ j = (1, 2, \dots, m)$.

This study aims to investigate the ability to detect anomalies in global financial markets through Graph Autoencoder (GAE), that is, an artificial neural network that receives input data through an encoder responsible for building a compact latent representation of the graph $G$ and a decoder that reconstructs the original graph from the compact latent model. To achieve this goal we used non-extensive entropy to measure the level of uncertainty in global financial markets before, during, and after a crisis period. 

The paper is organized as follows: section 2 presents the literature review conducted to identify gaps, state-of-the-art methods, and potentialities; Section 3 presents the methodology and data used, and describes the models and methodological strategies selected for evaluation; Section 4 presents the main results and discussions; and, finally, section 5 presents the conclusion and ideas for future research.

% Literature Review
\section{Background and Related Works}

The applications of GNN for anomaly detection in time series can be obtained through telemetry data, that is, data for monitoring the status of a particular system. The work proposed by \cite{XIE2021GNNAD} analyze anomaly detection in satellite systems with an accuracy of 98\%, demonstrating a high level of efficiency and robustness. Additionally, works such as \cite{LI2021DINADFIN} present graph-based strategies for detecting anomalies based on spurious relationships. The work of \cite{MOHIUDDIN2017ADFIN} apply a set of anomaly detection methods to the Australian Security Exchange (ASX), and the experimental results suggest that the LOF (Local Outlier Factor) and CMGOS (Clustering-based Multivariate Gaussian Outlier Score) are the best-performing anomaly detection techniques.

A graph neural network (GNN) refers to any artificial neural network that receives graph data as input \cite{WU2022GNN} and this approach is effectively applied in anomaly detection task \cite{WANG2022GNNAD}. According to \cite{WANG2022GNNFIN}, the challenges for applying GNN to financial data are the need for more transparent results (explainability), diversification in the type of task where GNN are applied considering that node-level tasks have greater prominence than edge-level and graph-level tasks, the construction of benchmark datasets that enable the reproduction of generated results, and, finally, the need for GNN-based systems for financial applications to have a scalable infrastructure (enabling applications for real-world problems).

Tsallis entropy is used in studies that evaluate financial markets. The study proposed by \cite{wang2018analysis} presents an approach called multiscale cross-distribution entropy based on Tsallis entropy, and the results indicate better performance in providing information about the relationship between two assets than the multiscale cross-sample entropy method \cite{yi2014multiscale}.

The aim of this study is to contribute to the literature on using of GNN for anomaly detection in global financial markets using Tsallis entropy as an anomaly score. This section presents the methodology applied in this work.

%\begin{figure*}[t]
%    \centering
%    \includegraphics[scale=0.15]{figures/framework.pdf}
%    \caption{Methodology framework}
%    \label{fig:methodology}
%\end{figure*}

%------
\subsection{Network Structure of Global Financial Markets}
%----- 

\subsubsection{Global Financial Market Graph}

The global financial market is formed by the relationships between assets that belong to each local financial market. In turn, each local financial market is a set of assets that are traded between companies seeking capital raising and investors. We assume that any asset that belongs to a particular local financial market is also part of the global financial market. The graph notation of a global financial market is 

\begin{equation}
	\mathcal{G}=(\mathcal{K}_i, \mathcal{W}_j),\forall i=(1,\dots,k)~\land~j=(1,\dots,w)
\end{equation}

\noindent where $\mathcal{G}$ is the global financial market, $\mathcal{K}$ is a set of all assets with $k\in\mathcal{K}$ represent the total number of assets in $\mathcal{G}$, $\mathcal{W}_{j}$ is a set of all asset edges with $w\in\mathcal{W}$ represent the total number of edges in $\mathcal{G}$.

%----- 
\subsubsection{Adjacency Matrix by Correlation Coefficients}

A data structure used in graph theory is the adjacency matrix. The adjacency matrix (Eq. \ref{eq:adj_matrix}) is an $n\times n$ matrix, where $n$ is the number of nodes in the graph, and each entry $a_{ij}$ shows whether there is an edge between two nodes or not.

\begin{equation}
	\label{eq:adj_matrix}
	A_{nn} = 
	\left| 
	\begin{array}{ccc}
		a_{11} & \dots & a_{1n} \\
		\vdots & \ddots & \vdots \\
		a_{n1} & \dots &  a_{nn}
	\end{array} 
	\right|\qquad a_{ij} = \left\{ \begin{array}{ll}
		0 & ~if~\nexists~(u, v)\\
		1 & ~if~\exists~(u,v)\end{array} 
	\right.
\end{equation}

The correlation matrix between the assets in $\mathcal{K}$ is a complete graph $\mathcal{G}$, where each element of $\mathcal{W}$ is a correlation. Thus, the correlation between two assets $u$ and $v$ represents the average of the products of their standardized values.

\begin{equation}
	cor(u,v)=\frac{cov(u,v)}{std(u)\cdot std(v)}
\end{equation}

\noindent where the covariance $(u,v)$ is the mean product of centered values, that is

\begin{equation}
	cov(u,v)=\frac{\sum_{i=1}^{p} (u_{i}-\bar{u})(v_{i}-\bar{v})}{p}
\end{equation}

Thus, the correlation matrix for the assets that belong to the global financial market was used to construct $\mathcal{G}$. The matrix $S_{kk}$ is a $k\times k$ matrix, for all $k\in\mathcal{K}$, where each entry $a_{kk}$ is defined by a threshold value ($\tau$).

\begin{equation}
	\label{eq:fin_matrix}
	S_{kk} = 
	\left| 
	\begin{array}{ccc}
		s_{11} & \dots & s_{1k} \\
		\vdots & \ddots & \vdots \\
		s_{k1} & \dots &  s_{kk}
	\end{array} 
	\right|\qquad s_{kk} = 
	\left\{ 
	\begin{array}{ll}
		0 & ~if~corr(u, v) < \tau\\
		1 & ~if~corr(u,v) \geq \tau
	\end{array} 
	\right.
\end{equation}

%----- 
\subsubsection{Estimated $\tau$ and Minimum Spanning Tree}

The approach to estimating $\tau$ was a \textit{winner-take-all} strategy in which only correlation coefficients above a limit are considered as edges. In this study, the 99th percentile was adopted as the limit, meaning that the final graph considers only the top 1\% of correlations, resulting in the graph $\mathcal{G}^{'}$. Finally, a minimum spanning tree (MST) algorithm was applied to $\mathcal{G}^{'}$ to verify the stability degree, as proposed in \cite{MICCICHE2003MSTFIN}.

%------
\subsection{Anomaly Detection with Graph Autoencoder}

\subsubsection{Graph Autoencoder} 

An autoencoder can be defined as a type of neural network algorithm that utilizes a compressed latent feature representation to learn and reconstruct input data. By minimizing the difference between the original input and the reconstructed output, an autoencoder is able to effectively extract and encode meaningful features from complex data sets. Figure 1 shows the architecture of a typical autoencoder, which consists of an encoder network that maps the adjacency matrix to a compressed latent feature representation, and a decoder network that maps the latent representation back to the original input space (reconstructed input).

To detect anomalies in global financial markets, the autoencoder maps the adjacency matrix to a low-dimensional latent space, where anomalies correspond to data points that deviate significantly from the learned patterns of normal data.

\begin{figure}[H]
	\centering
	\includegraphics[scale=0.25]{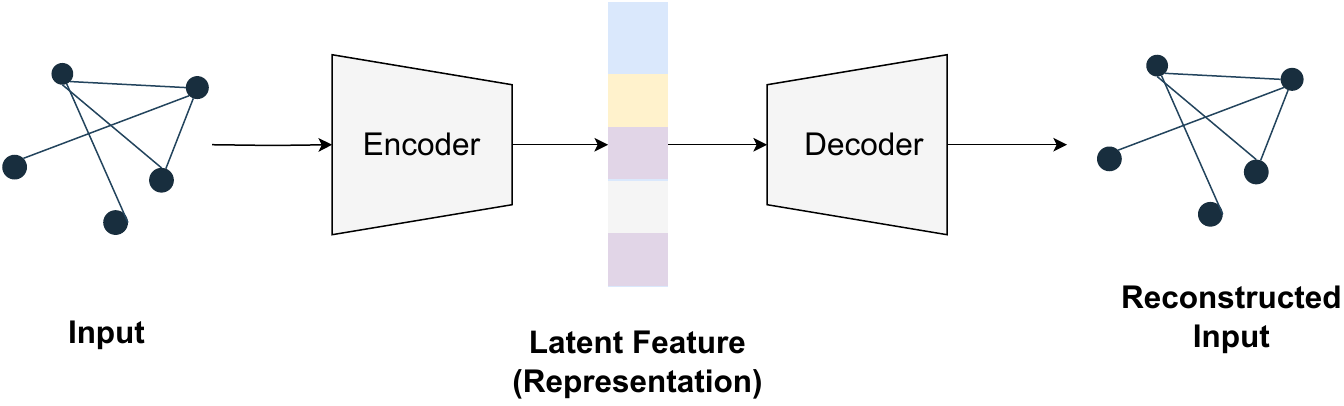}
	\caption{Autoencoder architecture}
	\label{fig:autoencoder}
\end{figure}

Considering a train set $\mathcal{D} = \{x_{i}|i=1,\dots,T\}, ~~\forall x_{i}\in \mathbb{R},~i\in \mathbb{N}$. The encoder can be written as

\begin{equation}
	h_{i} = g(x_{i})
\end{equation}

\noindent where the latent feature representation $h_{i}$ is the output of the encoder function $g$. And the decoder can be represented as 

\begin{equation}
	\tilde{x}_{i} = f(g_{x_{i}})
\end{equation}

\noindent where $\tilde{x}_{i}$ is the output of decoder function$f$. In turn, the training of an autoencoder is to find the functions $f$ and $g$ such that

\begin{equation}
	\arg\min_{f,g} <\Delta(x_{i}, \tilde{x}_{i})>
\end{equation}

\noindent where $\Delta$ is a loss function that penalize the difference between $x_{i}$ and $\tilde{x}_{i}$; and the operator $<\cdot>$ is the average over all observations. The reconstruction error (RE) is a metric to evaluate the ability of the autoencoder to reconstruct $x_{i}$. In this study, the mean squared errors (MSE),

\begin{equation}
	RE \equiv MSE = \frac{1}{T}\sum_{i=1}^{T}|x_{i}-\tilde{x}_{i}|^{2}
\end{equation}

\subsubsection{Anomaly Detection Scores}

As described by \cite{renyi1961measures}, let $\mathcal{P}=(p_{1}, p_{2}, \dots,p_{n})$ be a finite discrete probability distribution; the amount of uncertainty (concerning the output of an experiment) of $\mathcal{P}$ is called entropy of the distribution $\mathcal{P}$. This uncertainty (or randomness) is usually measured by Shannon entropy \cite{shannon1948entropy}, that is

\begin{equation}
	H(\mathcal{P}) = \sum_{k=1}^{n}p_{k}log_{2}\frac{1}{p_{k}}
\end{equation}

When applied to issues related to the financial market, entropy-based tools can measure the level of randomness in the markets \cite{delgado2019quantifying}, analyzing and predicting the behavior of stocks \cite{MAASOUMI2002291, GU2017215}, as well as pointing to a contagion effect of financial market uncertainty on other economic variables \cite{AHN2019uncertainty}.  

In this study, Tsallis entropy \cite{tsallis1988possible} was applied to measure uncertainty in global financial markets. Tsallis entropy is defined as

\begin{equation}
	S_q = \frac{1-\sum_{i=1}^W p_i^q}{q-1}
\end{equation}

\noindent where $p_i$ is the probability of finding the system in the $i$-th state, $q$ is a parameter that determines the degree of non-extensivity of the entropy, and $W$ is the number of states in the system. When $q\rightarrow 1$, Tsallis entropy reduces to $H[\mathcal{P}]$.

Algorithm \ref{algo:create_graph} performs a set of basic operations considering a matrix $P$ with $k$ vectors of size $n$ as input. Assuming that $k\leq n$, we can assume that the complexity of calculating the mean and standard deviation is $O(kn)$. Computing the covariance matrix involves three steps: (i) calculating the mean of each variable, $O(kn)$; (ii) centering the data by subtracting the mean of each variable, $O(kn)$; and (iii) multiplying the centered matrix and its transpose, $O(k^{2}n)$. The calculation of the correlation matrix can be performed in $O(k^{2})$. Similarly, creating a graph from a list of edges takes time $O(k^{2})$. Thus, the algorithm for constructing an undirected graph for asset returns is $O(k^{2}n)$, i.e., the execution time grows linearly with the input size $n$ and quadratically with the number of variables $k$.

\begin{algorithm}[H]
\caption{Undirected graph for asset returns}
\begin{algorithmic}[1]
\STATE $means = \text{mean of each vector of } P$
\STATE $m = \text{size of mean}$
\STATE $stds = \text{standard deviation of } P$
\STATE $P_{cov} = \text{compute the covariance matrix of } P$
\STATE $P_{corr} = \text{create an empty matrix } m \times m$
\FOR{$i = 0$ \TO $n$}
\FOR{$j = 0$ \TO $m$}
\STATE $P_{corr}[i, j] = \frac{P_{cov}[i, j]}{stds[i] \times stds[j]}$
\STATE $edges = \text{stack the correlation matrix } P_{corr}$
\STATE $edges\_w\_self = \text{remove the self-correlations}$
\STATE $asset\_graph = \text{create a graph with } edges\_w\_self$
\ENDFOR
\ENDFOR
\STATE \textbf{return} $asset\_graph$
\end{algorithmic}
\label{algo:create_graph}
\end{algorithm}

Algorithm \ref{algo:detectgnn} is a simplified representation of the process of detecting anomalies in global financial markets using the autoencoder architecture and Tsallis entropy as an anomaly score. Disregarding the running time for training the autoencoder, the algorithm performs a set of basic operations for each node in the global financial markets. Thus, we can estimate that the anomaly detection algorithm is $O(n)$.

\begin{algorithm}[H]
\caption{Anomaly detection algorithm}
\label{algo:detectgnn}
\begin{algorithmic}[1]
\STATE $h_{i} = g(x_{i})$
\STATE $\tilde{x}_{i} = f(g_{x_{i}})$
\STATE $loss = \mathcal{L}(x_{i},\tilde{x}_{i})$
\FOR{$node~~in~~G.nodes:$}
\IF{anomaly\_score(node)$>\tau$}
\STATE $anomalies\_list.append(node)$
\ENDIF
\ENDFOR
\STATE return $anomalies\_list$
\end{algorithmic}
\end{algorithm}

% Results and Discussion
\section{Results and Discussion}

This study uses the financial asset return dataset developed in \cite{KOSHIYAMA2021QUANTNET}. After a preliminary data cleaning to remove observations with different periods, the analysis considered a dataset with 1847 observations for 802 stocks. The data covers the period from 2004-10-27 to 2019-03-15.

The results were analyzed considering the global financial crisis (GFC) between 2007 and 2010. Thus, we follow three periods: the period before the crisis is from 2004-10-27 to 2007-06-27, the period during the crisis is from 2007-06-29 to 2010-08-24, and the period after the crisis is from 2010-08-25 to 2019-03-15. 

%Figure \ref{fig:corr_dist} report the distribution of correlation between stock returns in these three periods. Is possible to observe that before crisis the distribution of correlation for stock returns $\theta \sim \mathcal{N}(0, \sigma^{2})$. During and after crisis $\theta$ approximates another kind of distribution with a tail showing an increase in correlations.

%\begin{figure}[H]
%    \centering
%    \includegraphics[scale=0.45]{figures/correlation_distribution.pdf}
%    \caption{Distribution of correlations for assets returns before crisis, during crisis, and after crisis. Normal distribution fit: $N(0, \sigma^{2})$}
%    \label{fig:corr_dist}
%\end{figure}

\begin{figure}[H]
	\centering
	\includegraphics[scale = 0.2]{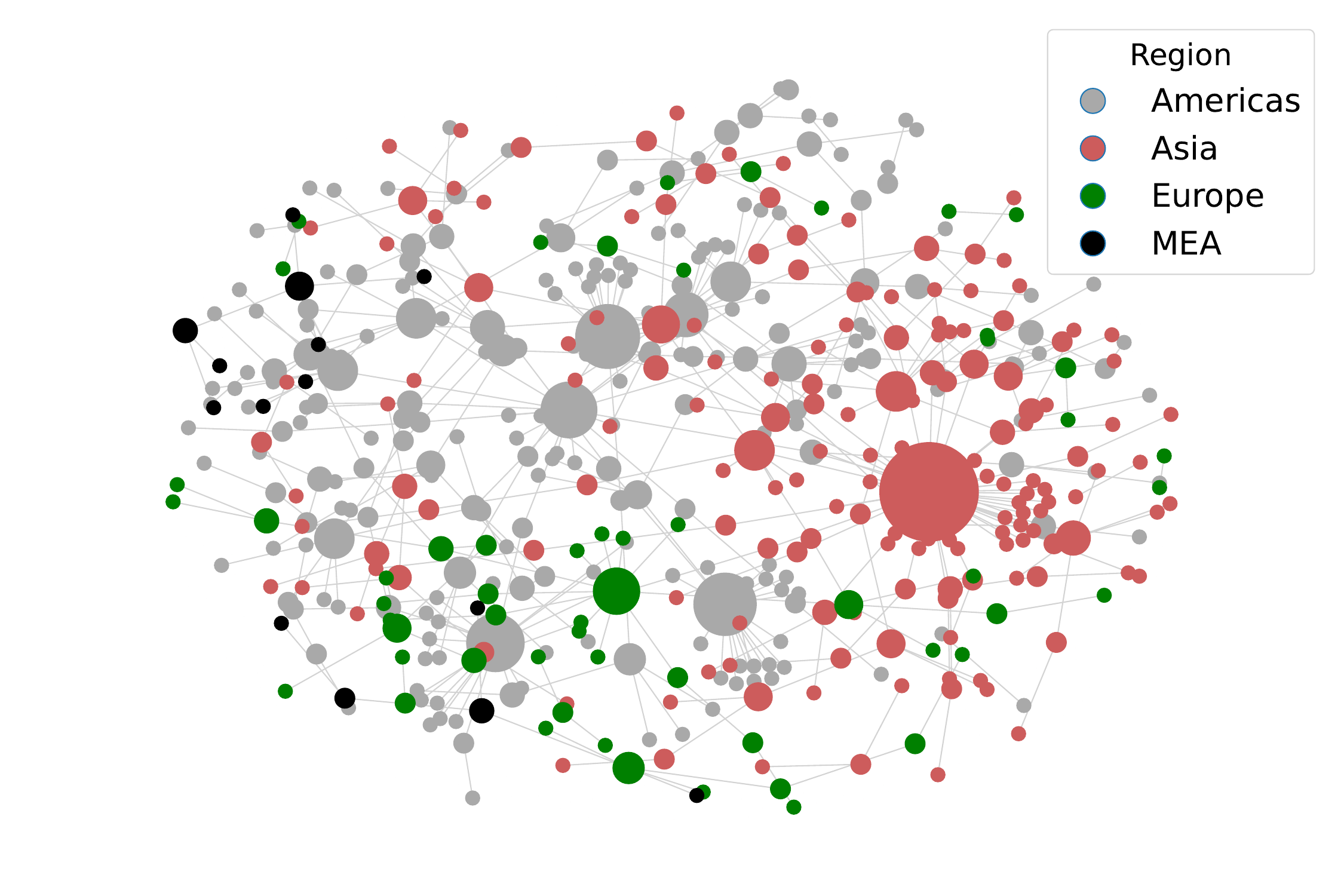}
	\caption{\textit{Before Crisis} - Association network and community structure in the high-correlated global financial market. The figure shows the global financial market correlation-based network with a threshold that selects only the high links weighted by correlations and runs in an MST algorithm.}
	\label{fig:graph_global_before}
\end{figure}

Figure \ref{fig:graph_global_before} shows the network for the global financial market before the crisis. It can be observed from Table \ref{tab:summary_results} that in this period, 455 edges were observed in the top 1\% of correlations. In this scenario, approximately 40\% of the stocks did not have edges. Analyzing the degree of centrality of each stock, it is possible to observe that the maximum value for the centrality degree is 60, with a distribution showing that most of the stocks are concentrated in the range between 0 and 40.

\begin{figure}[H]
	\centering
	\includegraphics[scale = 0.3]{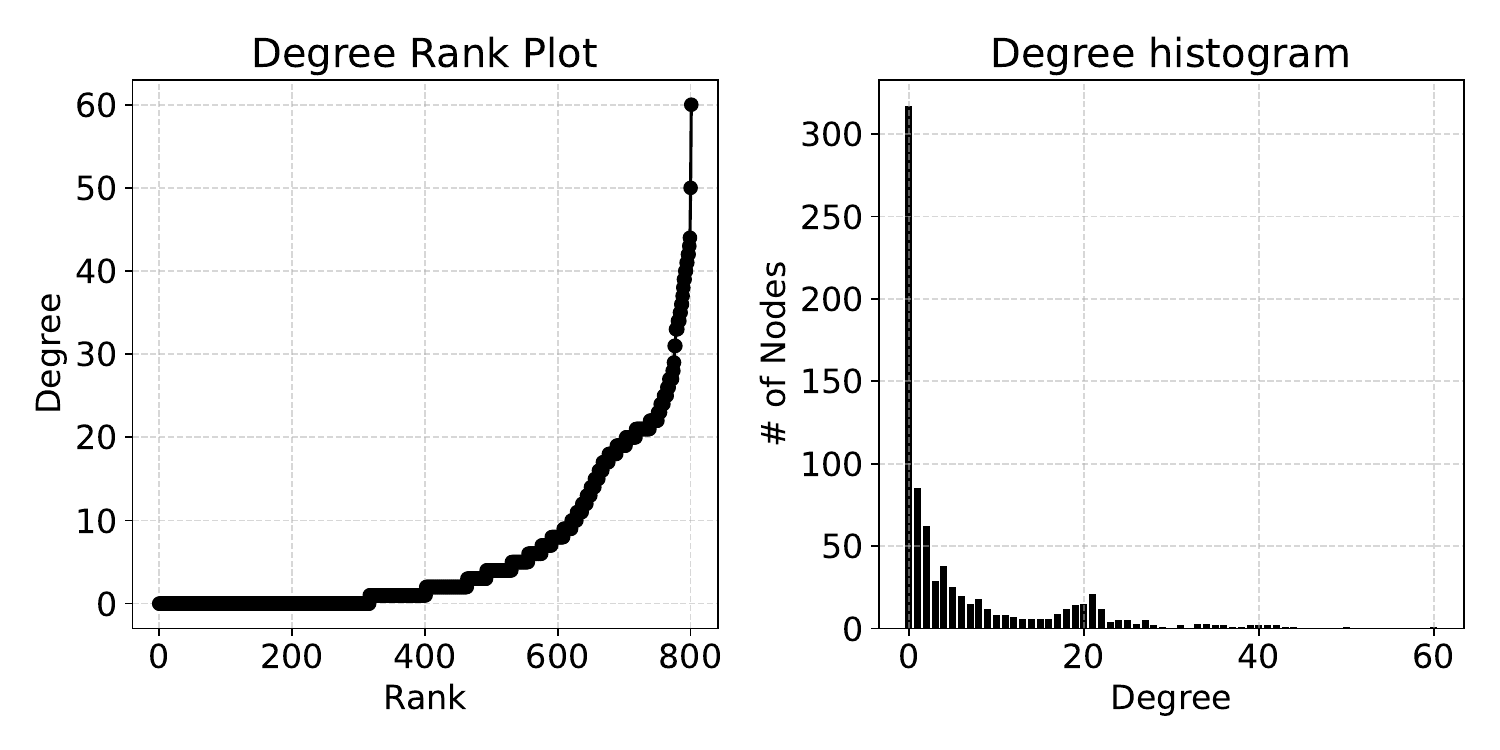}
	\caption{The rank of degree and distribution of degree centrality before crisis network}
	\label{fig:graph_global_before_dist}
\end{figure}

\begin{figure}[H]
	\centering
	\includegraphics[scale = 0.2]{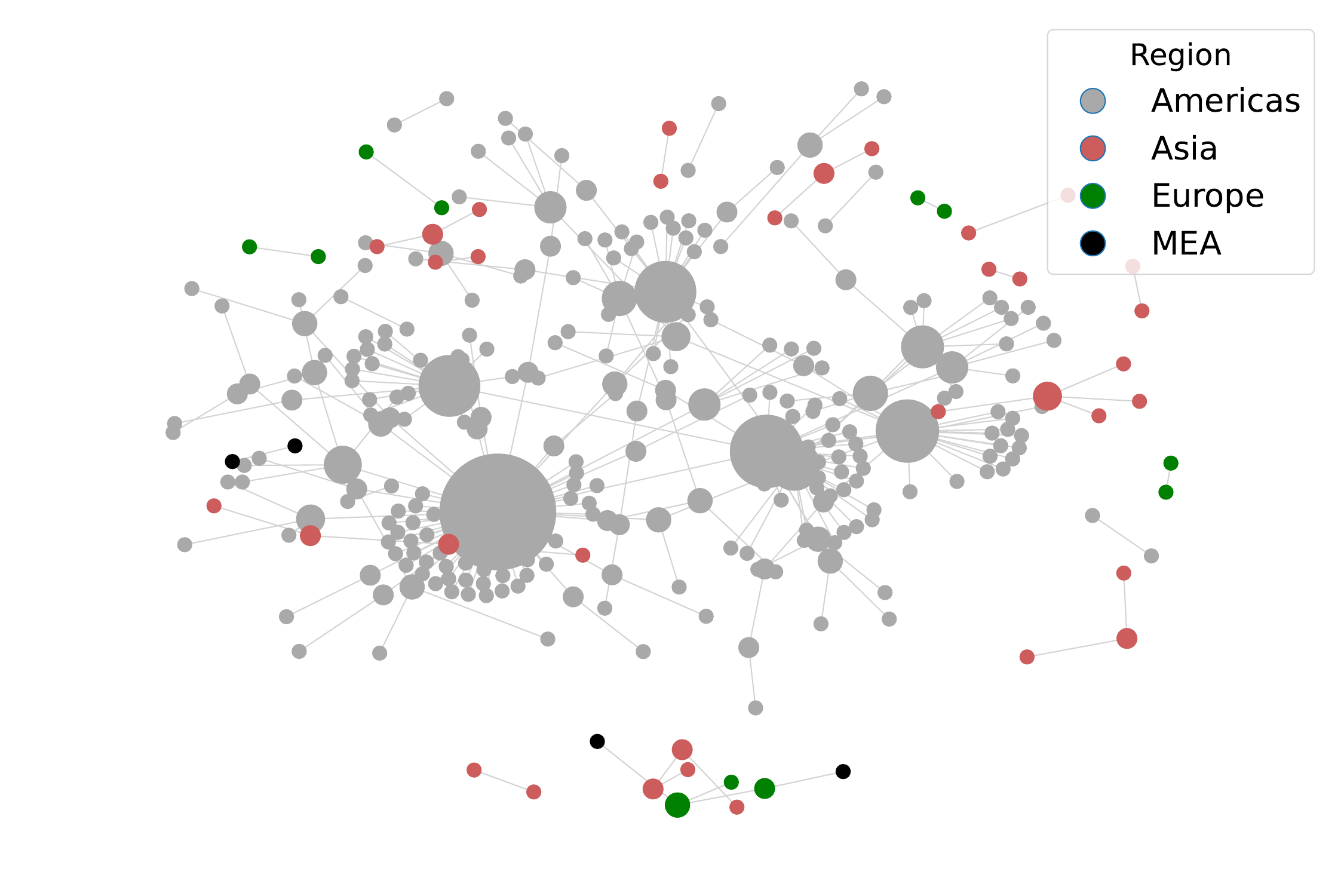}
	\caption{\textit{During Crisis} - Association network and community structure in the high-correlated global financial market. The figure shows the global financial market correlation-based network with a threshold that selects only the high links weighted by correlations and runs in an MST algorithm.}
	\label{fig:graph_global_during}
\end{figure}

\begin{figure}[H]
	\centering
	\includegraphics[scale = 0.3]{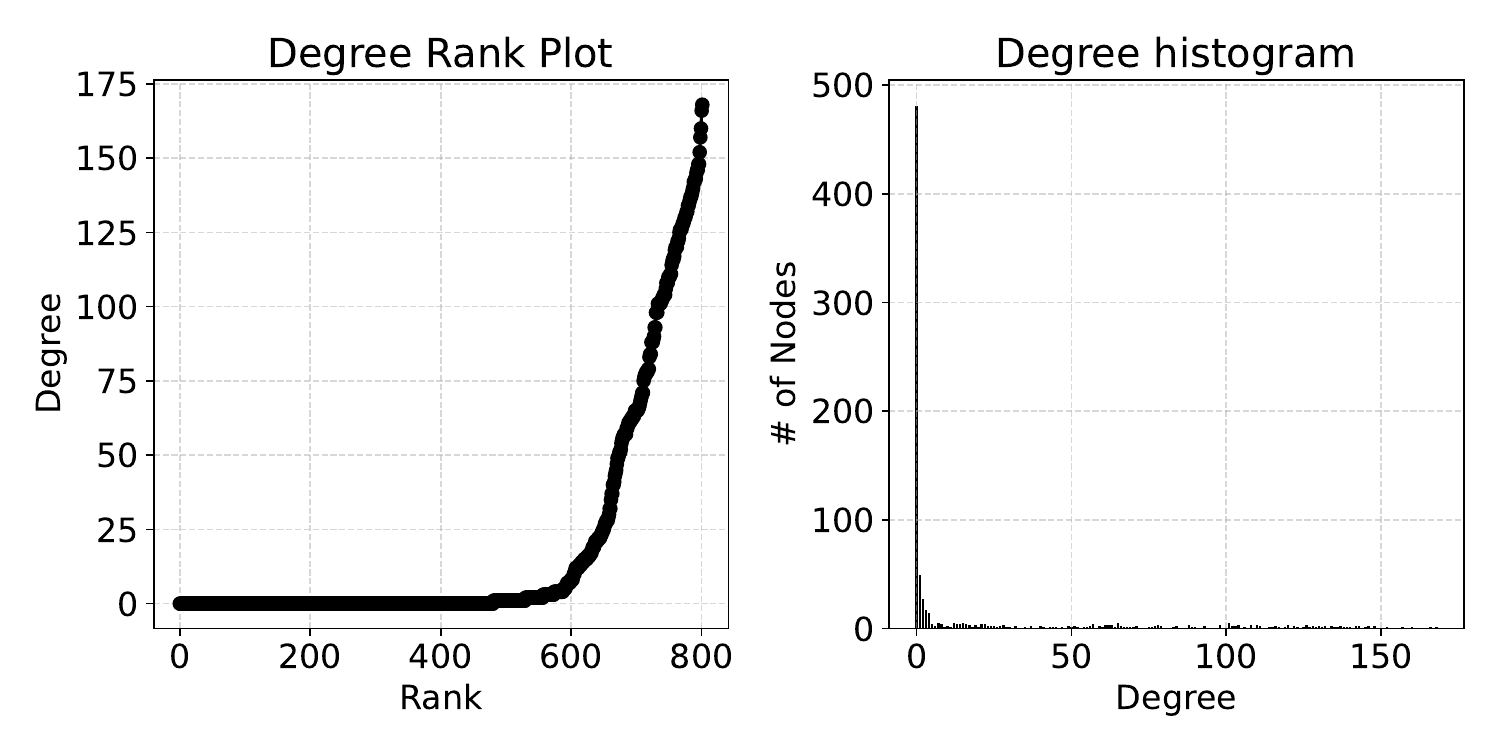}
	\caption{The rank of degree and distribution of degree centrality during crisis network}
	\label{fig:graph_global_during_dist}
\end{figure}

Figure \ref{fig:graph_global_during} presents the network for the global financial market during the crisis. In this period, 294 edges were observed in the top 1\% of correlations. In this scenario, approximately 64\% of the stocks did not have edges. Analyzing the degree of centrality of each stock, it is possible to observe that the maximum value for the degree of centrality is approximately 168, with a distribution showing that most stocks are concentrated close to 0.

\begin{figure}[H]
	\centering
	\includegraphics[scale = 0.2]{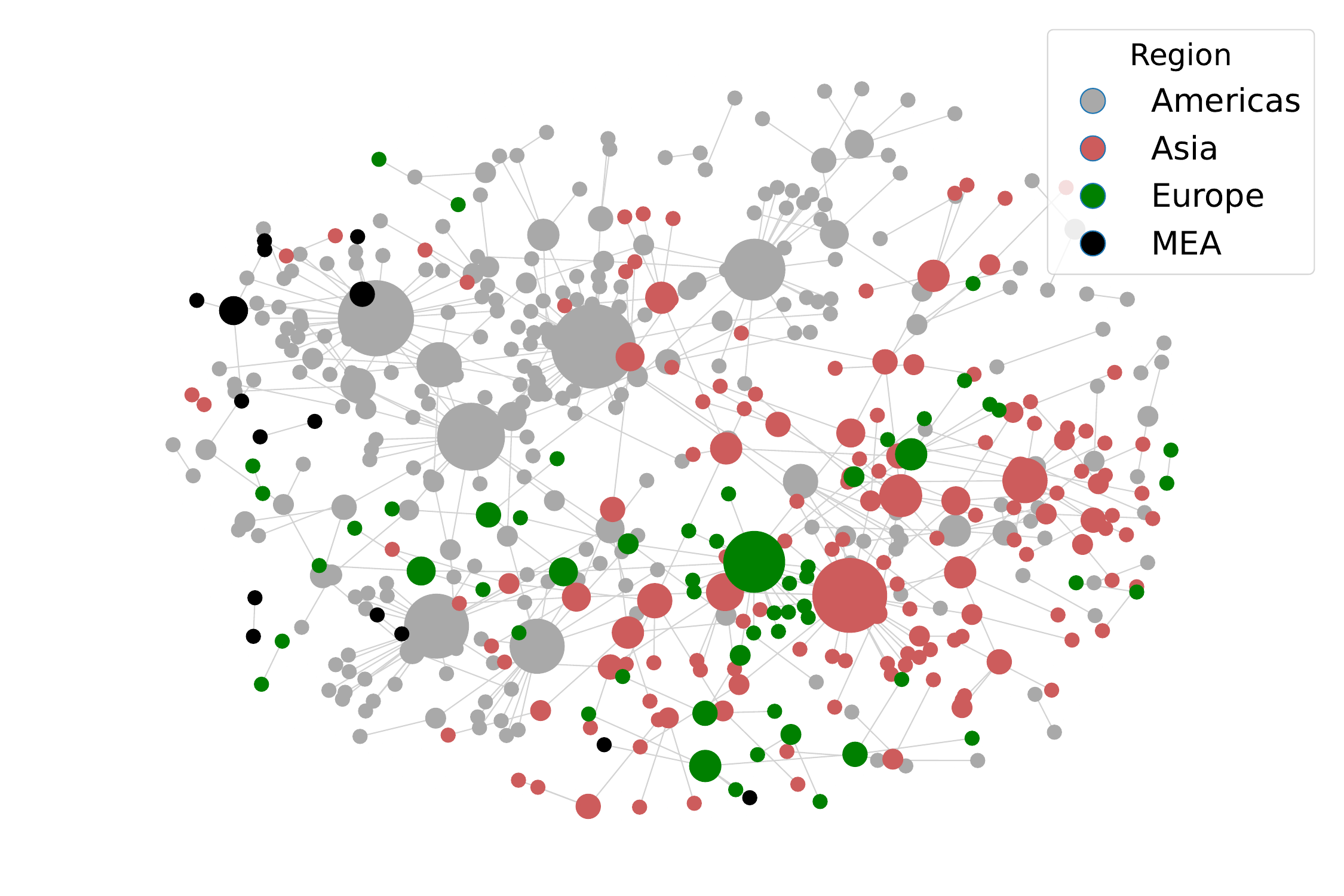}
	\caption{\textit{After Crisis} - Association network and community structure in the high-correlated global financial market. The figure shows the global financial market correlation-based network with a threshold that selects only the high links weighted by correlations and runs in an MST algorithm.}
	\label{fig:graph_global_after}
\end{figure}

Figure \ref{fig:graph_global_after} presents the network for the global financial market after the crisis. In this period, 488 edges were observed in the top 1\% of correlations. In this scenario, approximately 64\% of the stocks did not have edges. Analyzing the degree of centrality of each stock, it is possible to observe that the maximum value for the degree of centrality is 95, with a distribution showing that most stocks are concentrated in the range between 0 and 25.

\begin{figure}[H]
	\centering
	\includegraphics[scale = 0.3]{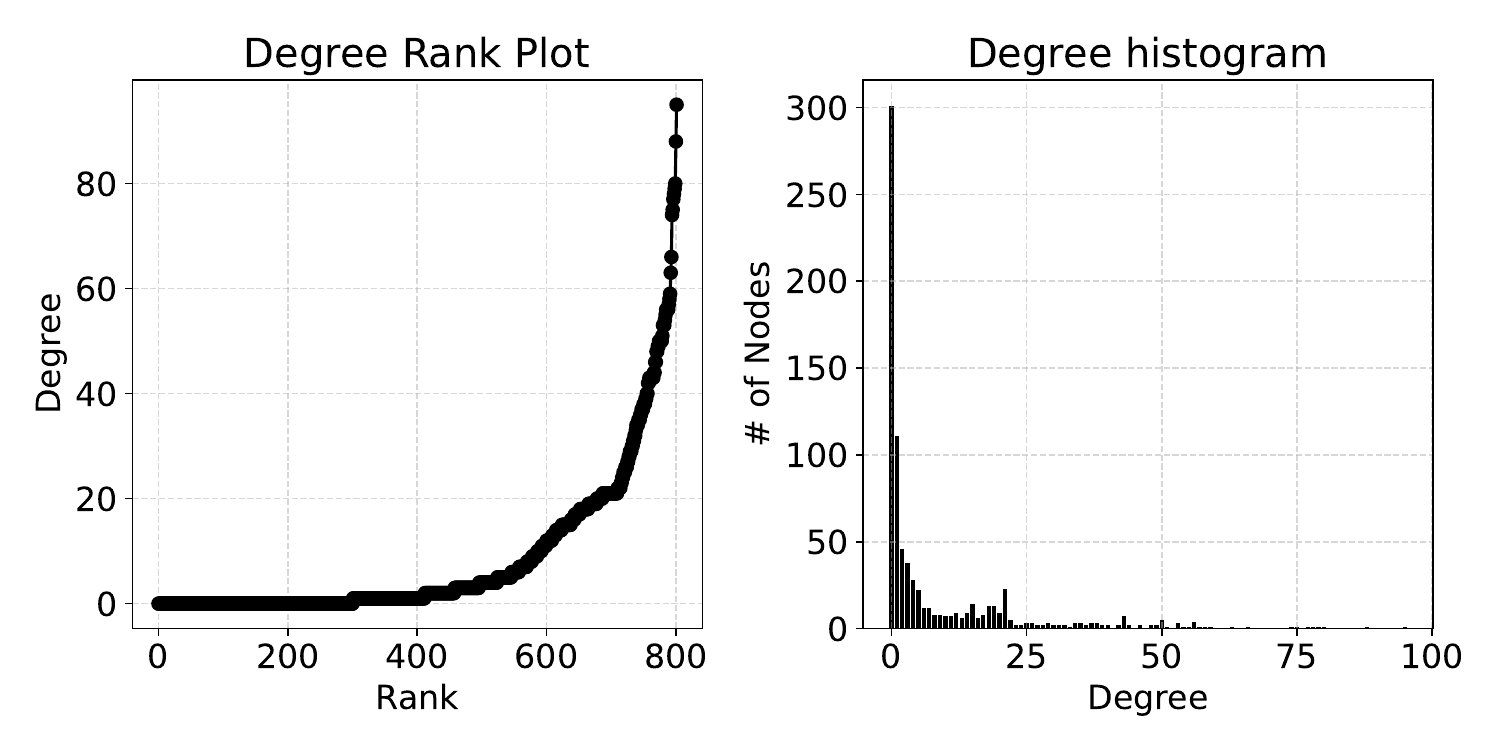}
	\caption{The rank of degree and distribution of degree centrality after crisis network}
	\label{fig:graph_global_after_dist}
\end{figure}

Through the autoencoder approach described in the methodology of this work, a convolutional neural network was trained with the task of label classification from a latent feature considering the correlation matrix for a set of countries that make up the selected financial markets. As observed in Table \ref{tab:summary_results}, the clustering coefficient shows that the tendency of data clustering decreases during the crisis period and increases after the crisis, following the same pattern observed through the number of edges for the top 1\% of correlations between assets. In other words, the results suggest that the financial markets become more interconnected and clustered after the crisis, which is consistent with the idea of increased global financial integration. Additionally, the analysis of the degree of centrality of each stock reveals that there are some stocks with high centrality that play an important role in the network, and that the distribution of centrality values is skewed towards lower values, indicating that most stocks are not highly connected.

\begin{table}[H]
\centering
\caption{Summary of results for before, during, and after crisis networks}
\label{tab:summary_results}
\begin{tabular}{lccc}
    \hline
    & \textbf{Before} & \textbf{During} & \textbf{After} \\\hline
    Number of Edges & 455 & 294 & 488 \\
    Nodes Without Edges & 39.52\% & 59.97\% & 37.53\% \\
    Max Degree & 60 & 168 & 95 \\
    Mean Degree & 6.14 & 18.38 & 8.49 \\
    Std Degree & 9.43 & 38.36 & 14.70 \\
    Clustering Coeff. & 0.30 & 0.26 & 0.37
    \\\hline
\end{tabular}
\end{table}

\begin{figure}[H]
	\centering
	\subfloat[\centering Before Crisis]{{\includegraphics[width=1.8cm]{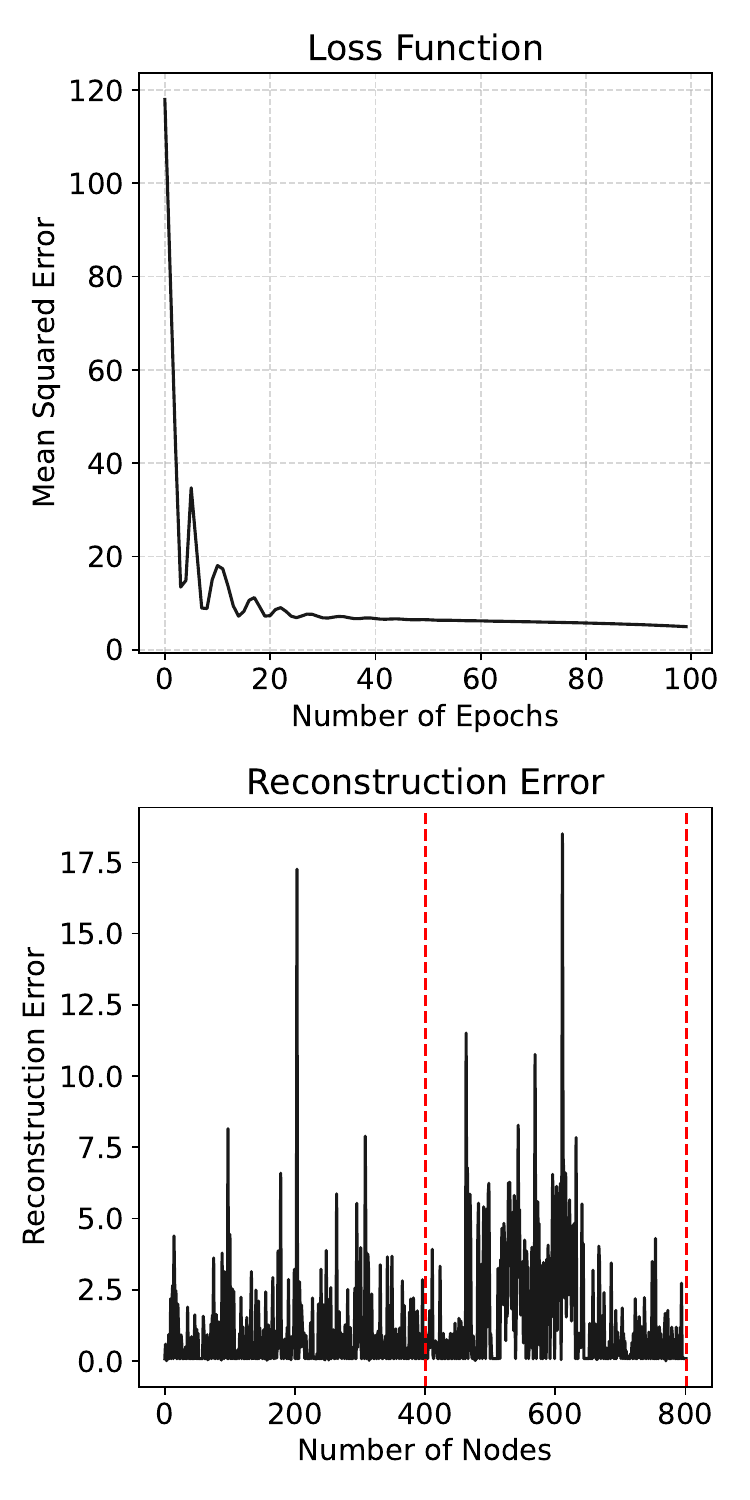} }}%
	\quad
	\subfloat[\centering During Crisis]{{\includegraphics[width=1.8cm]{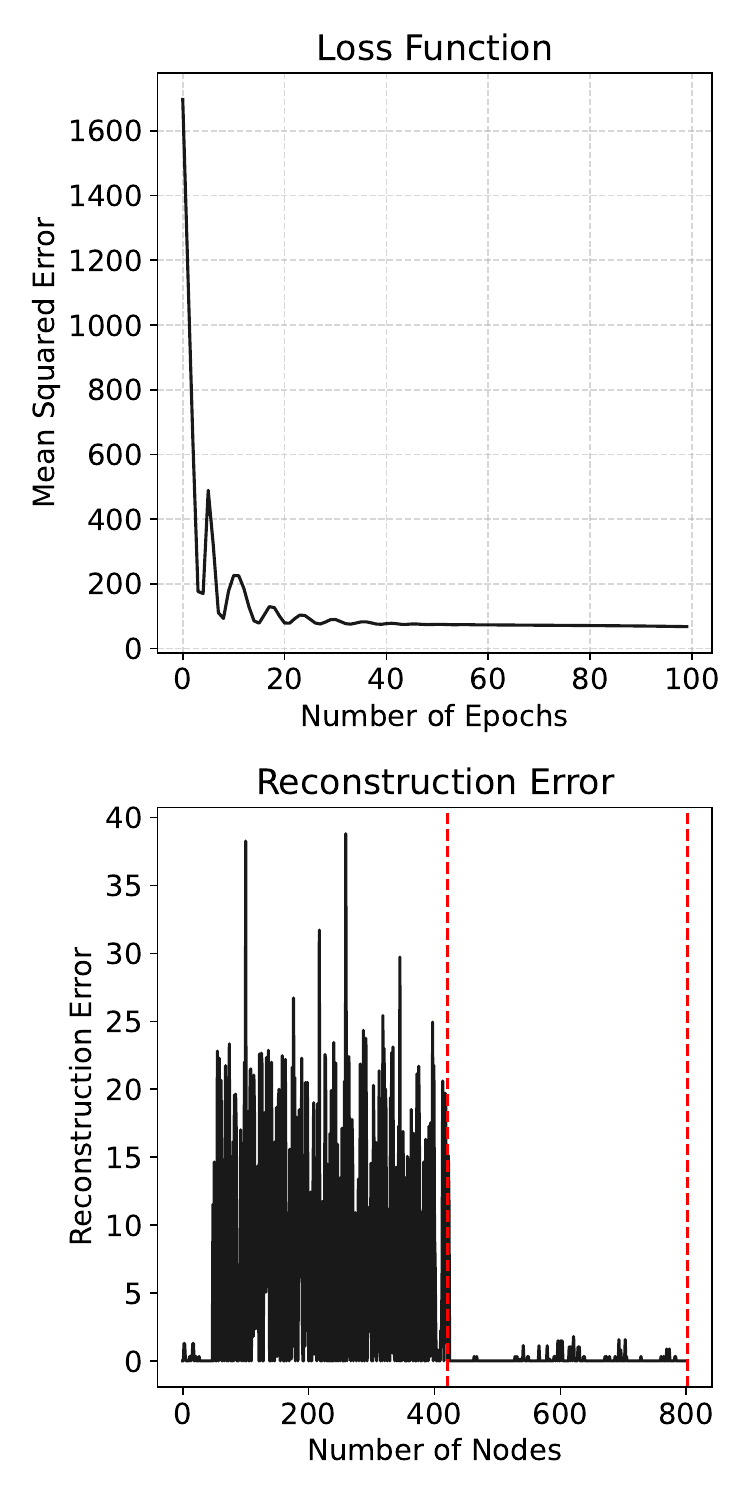} }}%
	\quad
	\subfloat[\centering After Crisis]{{\includegraphics[width=1.8cm]{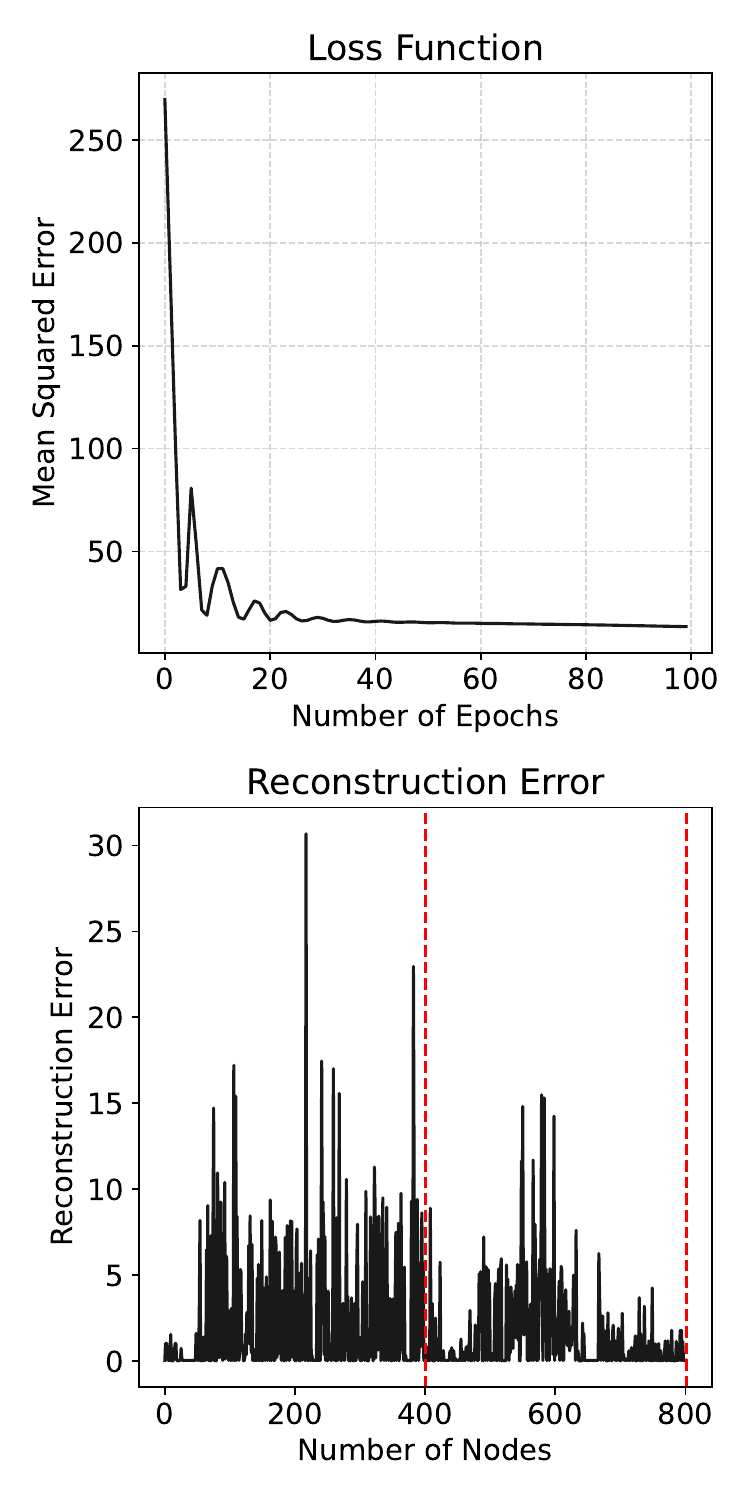} }}%
	\caption{Loss function and reconstruction error of graph neural network training}%
	\label{fig:training}%
\end{figure}

The Figure \ref{fig:training} presents the outcomes obtained by applying the loss function during the training process of the GNN. The results indicate that the model was successful in reducing its error for the training set in all three periods analyzed. This suggests that the GNN was capable of learning and improving its predictions over time. Furthermore, it are observed that the GNN had greater difficulty reconstructing the adjacency matrix before and after the crisis periods, in comparison to during the crisis period. This finding suggests that the model may have struggled to capture the underlying patterns and dynamics of the network during the pre- and post-crisis periods, which could have led to less accurate predictions of the network's behavior.

\begin{figure}[H]
	\centering
	\includegraphics[scale = 0.25]{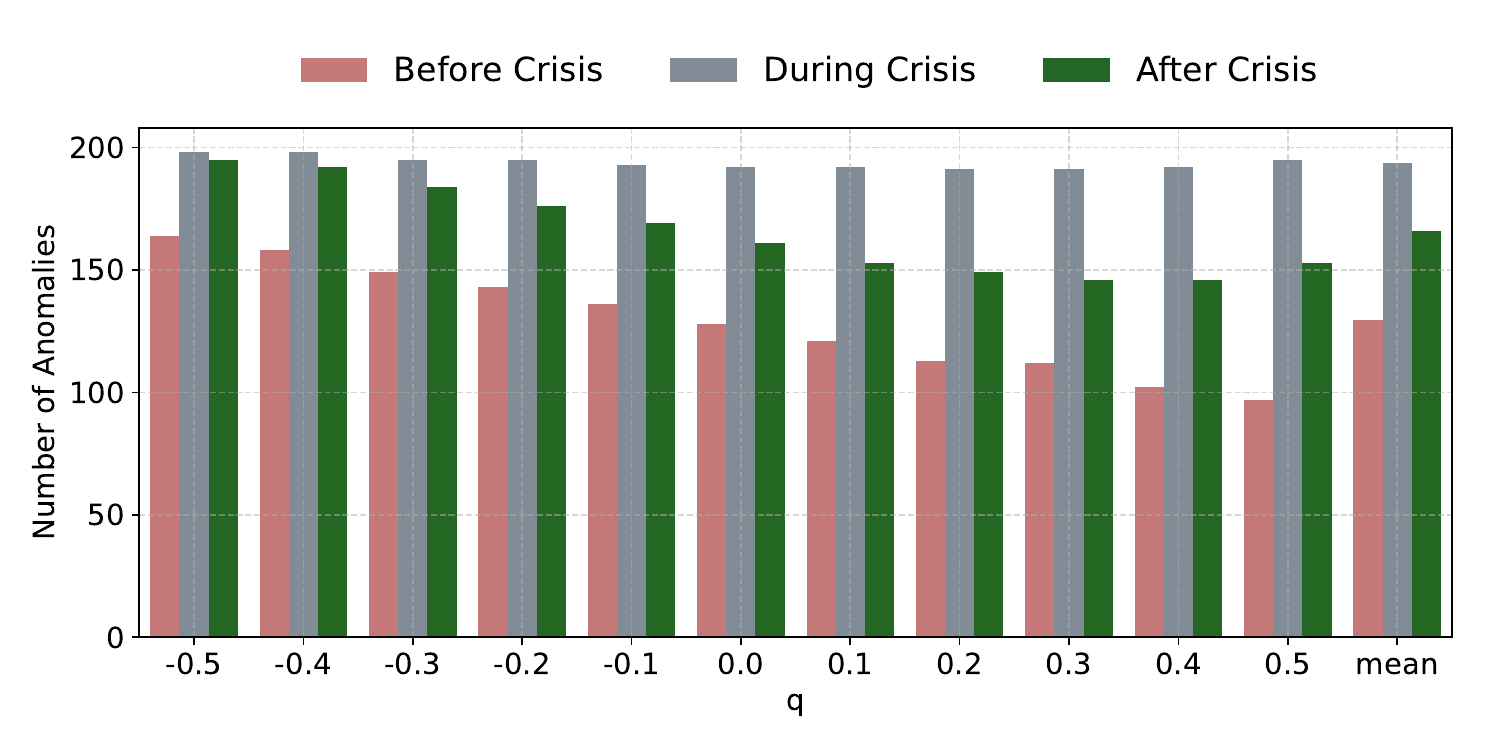}
	\caption{Number of anomalies by period}
	\label{fig:tsallis-score}
\end{figure}

Figure \ref{fig:tsallis-score} shows that, in the before-crisis period, the number of detected anomalies decreased as the value of $q$ increased. The same behavior is observed in the after-crisis period. However, the results indicate that, during the crisis period, there is stability in the number of detected anomalies, which becomes higher than the number observed in the other periods. This finding suggests that the parameter $q$ does not influence anomaly detection during the crisis period. To test the null hypothesis that the anomaly detection is different among the periods, a t-test was applied to compare the mean between two groups.

Table \ref{tab:results_test} shows the results of the t-test applied to the number of anomalies detected by Tsallis entropy for positive and negative values of $q$ (ranging from -0.5 to 0.5). Based on the analysis of the p-value, it is possible to reject the null hypothesis of equality between the means. Therefore, the results generated by the anomaly detection in the three periods are statistically different.

\begin{table}[H]
	\centering
	\caption{Summary results for T-test}
	\label{tab:results_test}
	\renewcommand{\arraystretch}{1.3}
	\begin{tabular}{lrr}
		\hline & t-statistic & p-value \\
		\hline
		Before Crisis vs During Crisis & -10.3308 & $6,65 \times 10^{-10}$ \\
		Before Crisis vs After Crisis & -4.5497 & $1.57 \times 10^{-4}$ \\
		During Crisis vs After Crisis & 5.4632 & $1.72 \times 10^{-5}$ \\ 
		\hline
	\end{tabular}
\end{table}

%\begin{figure}[H]
%    \centering
%    \includegraphics[scale = 0.28]{figures/tsallis_volatility_r2.pdf}
%    \caption{Caption}
%    \label{fig:tsallis_r2}
%\end{figure}

%\begin{figure*}[t]
%    \centering
%    \includegraphics[scale=0.5]%{figures/anomalies_tsallis_hist_before.pdf}
%\caption{Scatter plots between }
%    \label{fig:my_label}
%\end{figure*}

% Conclusion
\section{Conclusion}

The detection of anomalies in financial markets is an important task for different economic agents. Equity fund managers can use this information to improve their decision-making process and, eventually, improve their results. Central banks can use anomaly detection to identify dysfunctions in markets and act to mitigate the harmful effects of an economic crisis originating from the financial system, thus fulfilling their role as regulators and stabilizers of price levels, employment, and output.

The objective of this article was to investigate the use of graph neural networks for anomaly detection in a graph constructed from the correlation matrix for the stocks that make up different local financial markets, forming a representative graph of the global financial market. The results indicate a trade-off between the correlations observed during the crisis period: the correlation between all assets increases, but when we consider only the stocks with more connections (top 1\%), the correlation decreases. Thus, we indicate that during a crisis period, the graph for the global financial market becomes more sparse.

%This study also used nonextensive entropies as an anomaly score. When compared to mean absolute deviation (a traditional strategy for anomaly detection), nonextensive entropies show consistent results with an information gain, since data below a certain threshold are considered anomalous.

The limitations of the paper are primarily focused on the fact that the anomalies selected by the scores cannot be validated through a ground truth dataset. Therefore, in future studies, it is important to use a supervised learning strategy to calculate metrics that establish the accuracy of the model based on labeled anomalies.

\section{Acknowledgments}
The author thanks Simone D.J. Barbosa (DI/PUC-Rio) and Bernando Modenesi (University of Michigan) for their comments and suggestions. Any errors and inaccuracies are the responsibility of the author. 

\bibliographystyle{unsrtnat}
\bibliography{neurips}

\end{multicols}{2}
\end{document}